\PassOptionsToPackage{table}{xcolor}
\documentclass[letterpaper]{article}
\usepackage[T1]{fontenc}
\usepackage{iclr2027_conference,times}

\usepackage{amsmath,amsfonts,bm}

\def\eqref#1{equation~\ref{#1}}

\def\1{\bm{1}}

\DeclareMathAlphabet{\mathsfit}{\encodingdefault}{\sfdefault}{m}{sl}
\SetMathAlphabet{\mathsfit}{bold}{\encodingdefault}{\sfdefault}{bx}{n}

\usepackage{microtype,amsmath,amssymb,graphicx,booktabs,multirow}
\usepackage{xcolor,enumitem,xspace,url,etoolbox,wrapfig}
\usepackage[colorlinks=true,allcolors=blue]{hyperref}
\newcommand{\method}{RealtimeWAM\xspace}

\definecolor{realtimeblue}{RGB}{40,101,168}
\definecolor{tablehighlight}{HTML}{EEF5F8}
\colorlet{rankfirst}{tablehighlight}
\colorlet{ranksecond}{tablehighlight!50!white}
\colorlet{mainrankfirst}{rankfirst}
\colorlet{mainranksecond}{ranksecond}
\newcommand{\mainrankbest}[1]{\cellcolor{mainrankfirst}#1}
\newcommand{\mainrankrunnerup}[1]{\cellcolor{mainranksecond}#1}
\newcommand{\rankbest}[1]{\cellcolor{rankfirst}#1}
\newcommand{\rankrunnerup}[1]{\cellcolor{ranksecond}#1}

\title{RealtimeWAM: How Fast Can I Run\\My World Action Model?}
\hypersetup{pdftitle={RealtimeWAM: How Fast Can I Run My World Action Model?},pdfauthor={},pdfsubject={},pdfkeywords={}}
\author{\parbox{\textwidth}{\centering\normalfont
{\fontsize{10}{12}\selectfont\bfseries
Huanan Liu$^{1}$,\enspace Ye Li$^{1}$,\enspace Kangye Ji$^{1}$,\enspace Xiaoyu Chen$^{1}$\thanks{Work done during an internship at Tsinghua University.},\enspace Hanyun Cui$^{1}$,\enspace Yutian Shen$^{1}$\\
Yuan Meng$^{1}$,\enspace Chenglei Wu$^{2}$,\enspace Jingyan Jiang$^{1}$,\enspace Bo Li$^{3}$,\enspace Zhi Wang$^{1}$\thanks{Corresponding author.}\\[5pt]}
{\fontsize{9}{11}\selectfont
$^{1}$Tsinghua University\quad $^{2}$YuanxingGuangnian Robotics\quad $^{3}$Nanjing University}
}}
\hypersetup{pdfauthor={Huanan Liu, Ye Li, Kangye Ji, Xiaoyu Chen, Hanyun Cui, Yutian Shen, Yuan Meng, Chenglei Wu, Jingyan Jiang, Bo Li, Zhi Wang}}

\newcommand{\rrFullMatrixRows}{%
OpenWAM & 10 & 672.83 & 450.90 & 223.27 \\
OpenWAM & 4 & 240.53 & 205.49 & 101.95 \\
OpenWAM & 2 & 127.31 & 109.09 & 59.11 \\
\midrule
FastWAM & 10 & 214.33 & 99.68 & 33.52 \\
FastWAM & 4 & 104.61 & 50.27 & 27.99 \\
FastWAM & 2 & 71.55 & 38.69 & 21.56 \\
}

\newcommand{\rrNumericalRows}{%
OpenWAM & 10 & 3 & 0.001288 & 0.007812 \\
OpenWAM & 4 & 3 & 0.001245 & 0.011719 \\
OpenWAM & 2 & 3 & 0.002375 & 0.031250 \\
\midrule
FastWAM & 10 & 8 & 0.001449 & 0.015625 \\
FastWAM & 4 & 8 & 0.001384 & 0.015625 \\
FastWAM & 2 & 8 & 0.002515 & 0.031250 \\
}

\iclrfinalcopy
\makeatletter
\renewcommand{\@maketitle}{\vbox{\hsize\textwidth
  \hrule height 1.5pt
  \vskip 7pt
  {\centering\fontsize{16}{18}\selectfont\bfseries\@title\par}
  \vskip 7pt
  \hrule height 0.5pt
  \vskip 10pt
  {\centering\@author\par}
  \vskip 3pt
}}
\makeatother

\begin{document}
\maketitle
\fancyhead{}
\renewcommand{\headrulewidth}{0pt}
\begingroup
\setlength{\floatsep}{8pt plus 2pt minus 2pt}
\setlength{\textfloatsep}{10pt plus 2pt minus 2pt}
\setlength{\intextsep}{10pt plus 2pt minus 2pt}
\setlength{\abovecaptionskip}{6pt}
\begin{figure}[!ht]
\centering
\includegraphics[width=\textwidth]{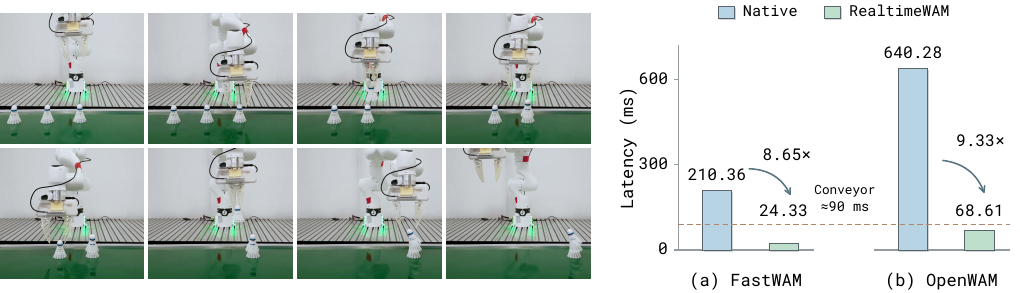}
\caption{\textbf{Responsive robot control with RealtimeWAM.}
Left: stacking three shuttlecocks on the conveyor requires rapid responses
to scene changes. Right: mean measured inference latency on this task
from Table~\ref{tab:real-world-tasks}, on RTX 4090. The dashed line marks the
empirically observed conveyor-task threshold of approximately 90\,ms,
above which task success was substantially lower
(Appendix~\ref{app:real-world-setup}).}
\label{fig:opening}
\end{figure}

\begin{abstract}
World Action Models (WAMs) combine visual dynamics modeling with action
generation, but their high inference latency limits responsive robot control.
Recent efforts accelerate inference by removing explicit future-video
generation at test time, as in FastWAM, an approach that requires a
specially tailored architectural design. More general caching
strategies exploit feature redundancy, but redundancy alone does not capture
the changing computational demands of closed-loop control. To address these
challenges, we present \method, a general, training-free framework that
coordinates parallel execution with adaptive computation for low-latency
inference across diverse WAM architectures. We exploit layerwise dependencies
to overlap observation processing with prediction. However, concurrent
branches still compete for GPU resources, limiting the benefit of parallel
execution. We therefore adapt computation throughout the pipeline through
selective reuse, caching observation features in visually stable regions and
reusing Transformer residuals while reserving additional refinement for
small predicted adjustments. We evaluate \method\ on FastWAM and OpenWAM across RoboTwin, LIBERO,
and LIBERO-Plus. On an RTX 4090, measured mean inference latencies
are 24.09 and 63.09\,ms, corresponding to average speedups of 8.90$\times$
and 10.67$\times$. Average success rates are 82.75\% and 87.41\%, respectively,
within 0.02 and 0.53 percentage points of native inference.
Across five real-world tasks, \method\ improves average success rates over
native inference by 17.2 and 37.2 percentage points on FastWAM and OpenWAM,
respectively.
Project page: \url{https://anonymous.4open.science/w/realtimewam/}.
\end{abstract}

\section{Introduction}
\setcounter{topnumber}{1}

World Action Models (WAMs) integrate visual dynamics modeling with action
generation, providing predictive representations for robotic control
\citep{cen2025worldvla,bi2025motus,kim2026cosmospolicy,yuan2026fastwam,wang2026openwam}.
However, their substantial inference cost limits how quickly a robot can
respond to changes in its environment. This limitation is particularly
problematic in dynamic manipulation, where the scene continues to evolve
while the model generates an action~\citep{black2025rtc}. For example, a moving object on a
conveyor may shift away from its observed position before the predicted
action is executed. Reducing latency from observation to action is therefore essential
for using WAMs to control robots in real time.

\begin{figure}[!t]
\centering
\includegraphics[width=\textwidth]{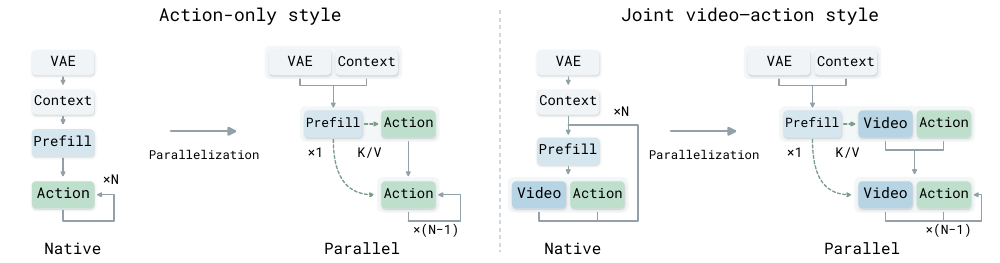}
\caption{\textbf{Parallel execution across WAM architectures.}
Left: action prediction alone. Right: joint video and action prediction.
Each pair contrasts native and parallel execution. Layerwise release of
observation K/V allows prediction to proceed alongside Prefill; shaded
regions group work available for concurrent execution.}
\label{fig:ffn-runtime}
\end{figure}

Recent work has reduced inference cost through architectural simplification
and computational reuse. For example, FastWAM removes future-video generation at inference while retaining
joint training with video
\citep{yuan2026fastwam}, and diffusion caching methods such as TeaCache
avoid repeated Transformer computation by reusing residual features
\citep{liu2025teacache}. These advances leave two further questions for
WAM execution in real time. First, reducing computation alone does not
resolve how observation processing and prediction should be scheduled
under their data dependencies and shared hardware constraints. Second,
caching decisions based on denoising similarity are not explicitly tied
to the responsiveness and precision required by the current action.
Fixed evaluation schedules likewise apply the same computational budget
throughout a rollout, despite changing execution requirements. Further
acceleration therefore calls for coordinating parallel execution with
computation that adapts to the current observation and predicted motion.

\begin{figure}[t]
\centering
\includegraphics[width=\textwidth]{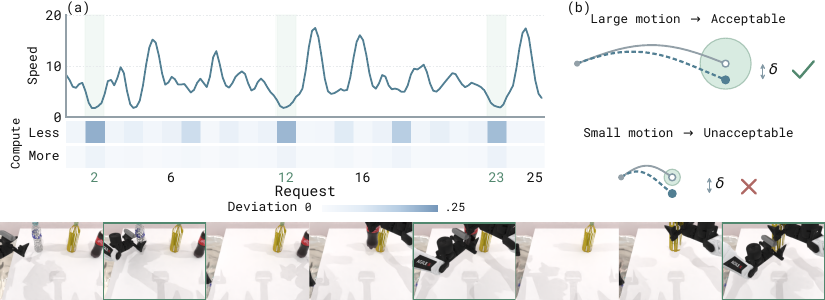}
\caption{\textbf{Changing motion and precision requirements.}
\textbf{(a)} Smoothed predicted speed (mm/control step) and relative deviation from a reference using more computation.
Darker cells indicate larger deviations; green marks identify selected observations.
\textbf{(b)} Illustrative cases: the same offset $\delta$ fits the large movement's wide tolerance but exceeds the small adjustment's narrow tolerance.
Measurement details are in Appendix~\ref{app:motion-motivation}.}
\label{fig:motivation}\label{fig:motion-consistency}
\end{figure}

To address these challenges, we present \method, a framework that coordinates parallel
execution with adaptive computation for existing WAM architectures
without additional training. We first exploit the layerwise dependencies
between processing the current observation (Prefill) and prediction. As each
layer produces the required observation features, the corresponding
prediction computation can begin while Prefill continues (Figure~\ref{fig:ffn-runtime}). This allows
Prefill to overlap with action prediction and, when present, future-video generation. However, concurrent branches compete for shared GPU compute
resources, limiting the speedup achievable through overlap alone.

To ease this contention, we exploit temporal redundancy across observations
to reduce the Prefill workload. Changes in the current observation guide
which tokens receive fresh computation, while cached features supply
the remaining positions. However, reducing
the number of refreshed tokens does not always reduce latency, as different
token counts can fall within the same latency plateau
(Table~\ref{tab:component-ablation}, right). We use this
available capacity to refresh additional tokens with larger feature changes
in the first layer, limiting reuse error at little extra latency. This reuse
policy accounts for hardware latency and reduces the workload competing
with prediction while keeping observation features current.

Token reuse reduces the cost of the first stage, but
subsequent denoising still requires repeated network evaluations. To further
accelerate inference, we adapt the refresh schedule of cached Transformer
residuals according to predicted motion. A coarse action prediction provides a
displacement signal for deciding whether further refinement is needed.
Motion amplitude and the effect of additional computation vary within an
episode (Figure~\ref{fig:motivation}), motivating a budget adapted to motion.
We allocate fewer evaluations to large predicted movements to favor rapid
replanning, while retaining additional evaluations for small adjustments.
We evaluate this allocation through closed-loop ablations in
Section~\ref{sec:motion-analysis} and paired prediction comparisons in
Appendix~\ref{app:motion-motivation}.
Refinement continues from the same initial denoising trajectory, preserving
the computation already performed.

We evaluate FastWAM and OpenWAM on RoboTwin, LIBERO, and LIBERO-Plus.
Combining execution optimizations with adaptive computation gives measured
average speedups of 8.90$\times$ and 10.67$\times$, respectively, relative
to native inference. Average success rates decrease by 0.02 and 0.53 percentage
points (Table~\ref{tab:main-comparison}). Controlled experiments examine
which scheduling and computation allocation choices account for these gains.

\begin{samepage}
Our main contributions are:
\begin{itemize}[label=$\bullet$]
\item We design a parallel schedule guided by layer dependencies that overlaps
observation processing with action prediction and, where applicable,
future-video generation.
\item We develop a token reuse policy that reduces the Prefill workload
and exploits hardware latency plateaus to refresh additional
informative tokens.
\item We adapt Transformer residual caching to predicted motion, using the displacement
of a coarse action prediction to determine whether further refinement is needed.
\item We evaluate two WAM architectures on three simulation benchmarks and five
real-world tasks, with controlled experiments examining execution, token
allocation, and predictive refinement.
\end{itemize}
\end{samepage}

\section{Related Work}

\paragraph{World action models and real-time inference.}
Diffusion Policy and $\pi_0$ generate actions with diffusion and flow
matching, respectively~\citep{chi2023diffusionpolicy,black2025pi0}.
WAMs couple visual prediction with action generation
\citep{shen2025videovla,bi2025motus,kim2026cosmospolicy,li2026lingbotva}.
FastWAM omits future-video generation at inference; OpenWAM jointly
predicts video and actions~\citep{yuan2026fastwam,wang2026openwam}.
Flash-WAM distills joint prediction to fewer steps~\citep{akbari2026flashwam}.
RTC, SmolVLA, and LingBot-VA overlap policy inference with action execution
\citep{black2025rtc,shukor2025smolvla,li2026lingbotva}.
Realtime-VLA and VLA-Perf examine inference optimization
\citep{ma2025realtimevla,jiang2026vlaperf}.
We overlap observation processing and prediction within one request.

\paragraph{Temporal reuse and diffusion caching.}
Diffusion caching reuses or extrapolates intermediate features across timesteps
\citep{ma2023deepcache,liu2025teacache,zou2024toca,liu2025taylorseer}.
FasterCache and PAB reuse diffusion features and attention outputs
\citep{lv2025fastercache,zhao2025pab}; TokenCache learns token importance
for selective caching~\citep{lou2024tokencache}.
PostDiff compares fewer denoising steps with mixed-resolution generation
and module caching~\citep{du2025postdiff}.
VLA-Cache, BAC, and EfficientVLA apply reuse at visual-token, block, and
diffusion-head levels~\citep{xu2025vlacache,ji2025bac,yang2025efficientvla}.
Our token reuse reduces Prefill contention and uses latency plateaus to
refresh additional tokens.

\paragraph{Adaptive computation.}
DynamicViT learns token sparsification; FastV prunes visual tokens using
attention scores~\citep{rao2021dynamicvit,chen2024fastv}.
PRANCE jointly adapts tokens and channels~\citep{li2025prance}.
AdaCache uses video motion to guide caching~\citep{kahatapitiya2024adacache}.
SAG learns observation-conditioned pruning; Test-time Sparsity combines
parallel inference with learned cached-feature selection
\citep{ji2026sag,ji2026testtimesparsity}.
SP-VLA couples scheduling with token pruning; ElegantVLA allocates
computation across perception, language, and action
\citep{li2025spvla,li2026elegantvla}.
TS-DP uses reinforcement learning to adapt speculative diffusion decoding
\citep{li2025ts}.
\method allocates Prefill and refinement budgets from observation changes
and predicted motion, without training a selector.

\section{Method}
\label{sec:method}
\setcounter{topnumber}{2}

\method coordinates execution order and fresh computation within each WAM
request (Figure~\ref{fig:ffn-layer-detail}). We first overlap independent
work using the model's dependencies. Hardware-aware token reuse then
reduces the observation workload competing with prediction, while
motion-adaptive refinement allocates subsequent network evaluations.

\begin{figure}[!t]
\centering
\includegraphics[width=\textwidth]{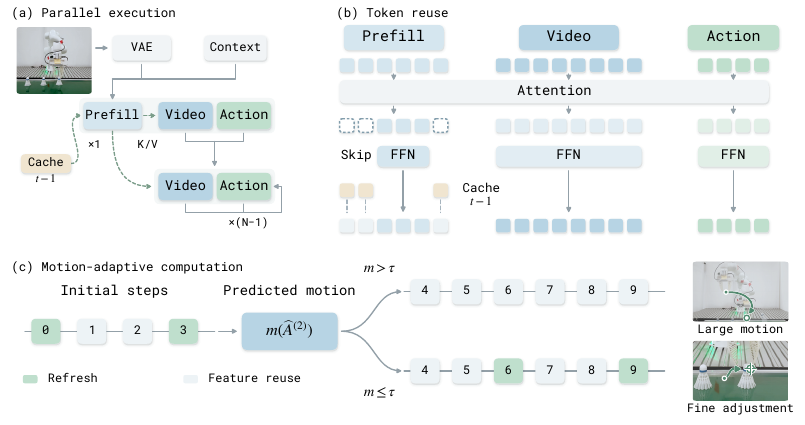}
\caption{\textbf{Coordinating token reuse and motion-adaptive refinement.}
\textbf{(a)} Observation processing proceeds alongside Video and Action prediction.
\textbf{(b)} Selected Prefill tokens receive fresh FFN outputs;
the remaining positions reuse cached outputs.
Future-video and action prediction execute their full branches at each selected
Transformer evaluation.
\textbf{(c)} Transformer refreshes at Euler updates 0 and 3 provide the
motion estimate that selects additional refreshes at 6 and 9.
Both schedules perform ten updates, with current input features and output heads
evaluated at every step.}
\label{fig:ffn-layer-detail}
\label{fig:adaptive-continuation}
\end{figure}

\subsection{Anatomy of WAM inference}
\label{sec:execution-limits}

At request $t$, a WAM maps images, robot state, and an instruction to an
$H$-step action chunk. Current-observation processing, termed \emph{Prefill},
provides features for action prediction and, when present, joint
video--action prediction. Under first-frame-causal attention and fixed
clean-frame conditioning, observation tokens are independent of the evolving
prediction. Their keys and values (K/V) provide the observation context
required by the predictive branches at each layer.

Let $R_t$ be the observation positions refreshed after the first layer,
and $N_t$ the number of full predictive Transformer evaluations. Request latency
can be decomposed as
\begin{equation}
 T_t \approx T_{\mathrm{first}}(R_t)+
 \sum_{j=2}^{N_t}T_{\mathrm{full},j}+
 T_{\mathrm{light}}+T_{\mathrm{finish}}.
 \label{eq:request-cost}
\end{equation}
The first stage includes input preparation, Prefill, and the initial prediction.
Thus $R_t$ controls the workload within the first stage,
while $N_t$ controls the amount of subsequent
refinement. $T_{\mathrm{light}}$ covers the remaining input embeddings,
output-head evaluations, and sampler updates; $T_{\mathrm{finish}}$ covers
final action postprocessing and return to the CPU. The measured request includes
online selection, data movement, and synchronization.

\subsection{Dependency-aware parallel execution}
\label{sec:parallel-execution}

We overlap independent input preparation: VAE encoding produces visual
latents while context preparation produces instruction features and
conditioning K/V, reusing instruction features when unchanged.

Prediction consumes observation K/V layer by layer. We release them immediately
after projection, allowing prediction to proceed alongside the remaining
Prefill attention and FFN operations (Figure~\ref{fig:ffn-runtime}).
This overlaps observation processing with the first predictive evaluation,
including the future-video branch when present.

Within each layer, local projections run concurrently; joint attention
waits for the required K/V, after which conditioning attention and FFNs
resume. Fused operators, preallocated buffers, and CUDA Graph replay
reduce transfers, repeated assembly, and launch overhead.

The overlapping branches share GPU resources, so we reduce their concurrent
workload by refreshing selected observation positions and reusing cached
outputs at the others.

\subsection{Hardware-aware token reuse}
\label{sec:prefill-reuse}

\begin{wrapfigure}{r}{0.48\textwidth}
\centering
\includegraphics[width=\linewidth]{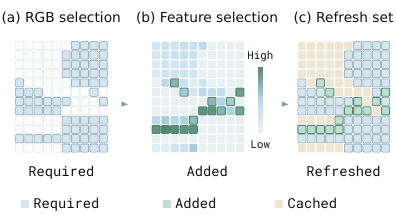}
\caption{\textbf{Prefill token allocation.}
\textbf{(a)} RGB change identifies required tokens.
\textbf{(b)} Feature change ranks the remaining tokens to fill spare capacity.
\textbf{(c)} Required and added tokens are refreshed; the rest stay cached.
Each cell is one observation token.}
\label{fig:bucket-allocation}
\end{wrapfigure}

The refresh budget must reflect observation changes and hardware cost.
GPU matrix operations exhibit tile- and wave-quantization effects
\citep{nvidiaMatmul}. Profiling our concurrent first stage reveals ranges
of token counts with nearly unchanged latency
(Table~\ref{tab:component-ablation}, right). We treat each plateau's upper
end as an execution capacity. Figure~\ref{fig:bucket-allocation} shows how
RGB changes establish required updates and feature changes guide the
additional updates that fill this capacity.

\paragraph{Selecting an execution capacity.}
For $M$ observation tokens, let $D_t$ contain positions whose mean absolute
RGB change exceeds $\delta$. We require these positions to refresh and
reserve a fraction $\alpha$ of the remaining positions to capture changes
not identified by RGB alone. Offline profiling of the concurrent first
stage gives execution capacities $\mathcal B$, including full refresh $M$. We choose
the smallest capacity covering the required and reserved updates:
\begin{equation}
 r_t=|D_t|+\lceil\alpha(M-|D_t|)\rceil,\qquad
 b_t=\min\{b\in\mathcal B:b\geq r_t\}.
 \label{eq:refresh-bucket}
\end{equation}
Each capacity uses a prepared execution shape for concurrent Prefill
and prediction.

\paragraph{Filling the selected capacity.}
The full first layer provides current features for every position. To
spend the remaining capacity on changing features, we rank positions
outside $D_t$ by normalized change from the preceding observation:
\begin{equation}
 s_{t,i}=\frac{2\|f^0_{t,i}-f^0_{t-1,i}\|_1}
 {\max(\|f^0_{t,i}\|_1+\|f^0_{t-1,i}\|_1,\epsilon)},\qquad
 R_t=D_t\cup\operatorname{Top}_{b_t-|D_t|}\{s_{t,i}:i\notin D_t\}.
 \label{eq:feature-fill}
\end{equation}
Here, $f^0_{t,i}$ is the first-layer feature and $\operatorname{Top}_k$
returns the $k$ highest-scoring indices. The set $R_t$ retains every
RGB-required position and supplements it with feature-ranked positions.
The same set serves subsequent layers; the feature reference advances
at every observation.

\paragraph{Executing fresh and cached tokens.}
For each reused layer $\ell>0$, selected positions update their FFN cache
entries while the others retain their last computed values. Writing
$\Phi_\ell$ for the FFN computation and $z^\ell_{t,i}$ for its
current input, the update is
\begin{equation}
 C^\ell_{t,i}=
 \begin{cases}
 \Phi_\ell(z^\ell_{t,i}),&i\in R_t,\\
 C^\ell_{t-1,i},&i\notin R_t.
 \end{cases}
 \label{eq:ffn-reuse}
\end{equation}
FFN reuse applies to Prefill; future-video and action tokens are fully computed
at each selected evaluation. Compatible operations batch refreshed
observation and prediction rows. Attention and current residual gates remain fully computed
(Appendix~\ref{app:prefill-contract}). Full refresh initializes the FFN cache;
episode or conditioning changes invalidate incompatible history.

\subsection{Motion-adaptive refinement}
\label{sec:adaptive-continuation}

Repeated Transformer evaluations remain on the critical path after the
first stage. We cache the input-to-output residual of the
Transformer. Each denoising step embeds the
current latent, adds the cached residual, and evaluates the output head
with the current timestep. Full evaluations refresh this residual.
Motivated by the varying precision demands in
Figure~\ref{fig:motivation}, we use predicted motion amplitude to allocate
refinement: large motions favor earlier replanning, while small adjustments
receive additional computation.

Both schedules share two initial Transformer refreshes. After the second,
we decode a coarse action prediction.
For current end-effector position $p_t$ and
predicted physical positions $\widehat p^{(2)}_{t,k}$, we use the maximum
displacement to choose the evaluation budget:
\begin{equation}
 m_t=\max_{1\leq k\leq H}\|\widehat p^{(2)}_{t,k}-p_t\|_2,
 \qquad N_t=2+2\,\mathbf 1[m_t\leq\tau].
 \label{eq:adaptive-continuation}
\end{equation}
The threshold $\tau$ is fixed during deployment. The 2F schedule refreshes
at Euler updates 0 and 3; the 4F schedule adds refreshes at 6 and 9.
Both perform ten Euler updates using the current output-head prediction
at each step. Refinement continues from the shared prefix with the same
noise and conditioning
(Figure~\ref{fig:adaptive-continuation}c).

Together, $R_t$ and $N_t$ allocate fresh computation within the first
stage and across later evaluations.
Section~\ref{sec:motion-analysis} compares fixed and adaptive budgets;
Appendix~\ref{app:rr-adaptive} gives the sampler settings and
paired-output diagnostics.

\section{Experiments}
\label{sec:experiments}

We evaluate \method on simulation and real-world manipulation tasks,
and ablate execution optimizations, token reuse, and motion-adaptive refinement.

\begingroup
\setlength{\floatsep}{8pt plus 2pt minus 2pt}
\setlength{\textfloatsep}{10pt plus 2pt minus 2pt}
\setlength{\intextsep}{8pt plus 2pt minus 2pt}
\setlength{\abovecaptionskip}{5pt}
\suppressfloats[t]
\begin{table}[t]

\caption{\textbf{Training-free acceleration across manipulation benchmarks.}
Columns report success rates (\%) and mean latency (ms); Average weights settings
equally. Caching baselines use compilation and CUDA Graph replay. Speedup
is original Native mean latency divided by method mean latency.
\colorbox{mainrankfirst}{Best} and \colorbox{mainranksecond}{second-best}
results exclude Native; lower latency is better.}
\label{tab:main-comparison}\label{tab:closed-loop}
\centering
\small
\newcommand{\mainboldtimes}{\scalebox{0.9}{\normalsize$\boldsymbol{\times}$}}
\setlength{\tabcolsep}{0pt}
\renewcommand{\arraystretch}{0.90}

\begin{tabular}{@{}p{54pt}p{57pt}*{4}{>{\centering\arraybackslash}p{39pt}}>{\centering\arraybackslash}p{35pt}>{\centering\arraybackslash}p{44pt}>{\centering\arraybackslash}p{43pt}@{}}
\toprule
& & \multicolumn{2}{c}{RoboTwin} & \multicolumn{2}{c}{LIBERO} & \multicolumn{3}{c}{Average} \\
\cmidrule(lr){3-4}\cmidrule(lr){5-6}\cmidrule(l){7-9}
Backbone & Method & Clean & Random & LIBERO & Plus & SR (\%) & Lat. (ms) & Speedup \\
\midrule
\multirow{6}{*}{\textbf{FastWAM}} & Native & 91.88 & 91.78 & 97.60 & 49.80 & 82.77 & 214.33 & 1.00$\times$ \\
\cmidrule(l){2-9}
 & VLA-Cache & 80.00 & 83.20 & 97.00 & 48.60 & 77.20 & 115.55 & 1.85$\times$ \\
 & TeaCache & 81.42 & \mainrankrunnerup{90.86} & 97.05 & 49.20 & 79.63 & \mainrankrunnerup{49.28} & \mainrankrunnerup{4.35$\times$} \\
 & ToCa & 91.35 & \mainrankbest{91.24} & 94.55 & 49.15 & 81.57 & 102.66 & 2.09$\times$ \\
 & BAC & \mainrankrunnerup{91.67} & 89.53 & \mainrankbest{97.50} & \mainrankbest{51.55} & \mainrankrunnerup{82.56} & 67.14 & 3.19$\times$ \\
 & \textbf{Ours} & \mainrankbest{\textbf{92.04}} & \textbf{90.50} & \mainrankrunnerup{\textbf{97.25}} & \mainrankrunnerup{\textbf{51.20}} & \mainrankbest{\textbf{82.75}} & \mainrankbest{\textbf{24.09}} & \mainrankbest{\textbf{8.90}\mainboldtimes} \\
\midrule
\multirow{6}{*}{\textbf{OpenWAM}} & Native & 93.64 & 94.06 & 99.30 & 64.75 & 87.94 & 672.83 & 1.00$\times$ \\
\cmidrule(l){2-9}
 & VLA-Cache & 93.36 & 89.94 & 98.40 & 62.70 & 86.10 & 400.26 & 1.68$\times$ \\
 & TeaCache & 90.42 & \mainrankbest{94.12} & 98.60 & 63.80 & 86.74 & \mainrankrunnerup{245.99} & \mainrankrunnerup{2.74$\times$} \\
 & ToCa & \mainrankrunnerup{93.42} & 93.98 & 98.10 & 63.70 & 87.30 & 440.42 & 1.53$\times$ \\
 & BAC & \mainrankbest{93.52} & 91.96 & \mainrankrunnerup{98.90} & \mainrankbest{66.70} & \mainrankbest{87.77} & 305.08 & 2.21$\times$ \\
 & \textbf{Ours} & \textbf{92.78} & \mainrankrunnerup{\textbf{94.02}} & \mainrankbest{\textbf{99.00}} & \mainrankrunnerup{\textbf{63.85}} & \mainrankrunnerup{\textbf{87.41}} & \mainrankbest{\textbf{63.09}} & \mainrankbest{\textbf{10.67}\mainboldtimes} \\
\bottomrule
\end{tabular}

\end{table}

\begin{table}[!t]

\caption{\textbf{Cumulative execution optimizations on RTX 4090 (ms).}
Values are mean latencies measured during complete closed-loop rollouts.
First: latency from prepared CPU inputs through Prefill and the first
predictive evaluation in a 2F run; Full 10F: request latency from a separate
10F run. Stages are cumulative. The plot shows first-forward
speedup over Native. Darker/lighter fills mark the lowest/second-lowest
latency among optimized stages.}
\label{tab:execution-stages}
\centering
\begin{minipage}[c]{0.72\linewidth}
\centering
\begingroup
\normalsize
\setlength{\tabcolsep}{0pt}
\renewcommand{\arraystretch}{1.0}
\small
\begin{tabular}{@{}>{\centering\arraybackslash}p{0.11\linewidth}p{0.25\linewidth}*{4}{>{\centering\arraybackslash}p{0.16\linewidth}}@{}}
\toprule
\multirow{2}{*}{Stage} & \multirow{2}{*}{Optimization} & \multicolumn{2}{c}{FastWAM} & \multicolumn{2}{c}{OpenWAM} \\
\cmidrule(lr){3-4}\cmidrule(l){5-6}
 & & First & Full 10F & First & Full 10F \\
\midrule
0 & Native & 67.24 & 214.33 & 81.37 & 672.83 \\
\cmidrule(l){2-6}
1 & + CUDA Graph & 36.83 & 104.09 & 69.34 & 524.69 \\
2 & + Compile & 36.21 & 99.68 & 60.62 & 450.90 \\
3 & + Parallel & \rankrunnerup{17.77} & \rankrunnerup{33.52} & \rankrunnerup{37.12} & \rankrunnerup{223.27} \\
4 & + Token reuse & \rankbest{17.69} & \rankbest{29.47} & \rankbest{31.62} & \rankbest{217.24} \\
\bottomrule
\end{tabular}
\endgroup
\end{minipage}\hspace{0.02\linewidth}%
\begin{minipage}[c]{0.26\linewidth}
\centering
\includegraphics[width=\linewidth]{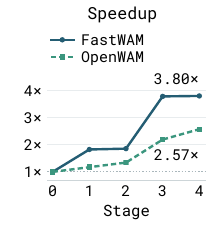}
\end{minipage}

\end{table}

\subsection{Experimental setup}

\textbf{Backbones, benchmarks, and baselines.}
We evaluate FastWAM and OpenWAM on RoboTwin 2.0 clean/randomized
\citep{chen2025robotwin2}, LIBERO~\citep{liu2023libero}, and
LIBERO-Plus~\citep{fei2026liberoplus}. Each backbone's native inference is the reference;
VLA-Cache~\citep{xu2025vlacache}, TeaCache~\citep{liu2025teacache},
ToCa~\citep{zou2024toca}, and BAC~\citep{ji2025bac} provide training-free
caching baselines. We enable compilation and CUDA Graph replay for the
caching baselines and \method. Baselines retain their original execution
schedules; \method adds the proposed parallel execution.
Table~\ref{tab:main-comparison} reports success rates per
setting and equally weighted averages over the four settings.
LIBERO category results appear in Appendix~\ref{app:benchmark-details}.

\textbf{Metrics and timing.}
We report success rate and full-request latency, with speedup relative
to native inference. Runtime measurements use BF16, batch size one, and a
modified 48\,GiB RTX 4090. Timing covers the prepared CPU image/state interface
through denormalized CPU actions, including input processing, model and
sampler execution, transfers, and synchronization. Compilation, graph
capture, warmup, profiling, and robot execution are excluded.
2F/4F denote two/four full Transformer evaluations within ten Euler updates.
The main comparison and execution ablations report mean latencies
measured during complete closed-loop rollouts.
Model-specific interfaces and checks appear in Appendix~\ref{app:rr-protocol}.

\subsection{Overall comparison}
\label{sec:runtime-results}

\textbf{Success--latency tradeoff.}
\method has the lowest reported average latency on both backbones
(Table~\ref{tab:main-comparison}). FastWAM latency decreases from
214.33 to 24.09\,ms (8.90$\times$), while the success rate changes from
82.77\% to 82.75\%. OpenWAM latency decreases from 672.83 to
63.09\,ms (10.67$\times$), with the success rate changing from 87.94\% to
87.41\%. Relative to Native with compilation and CUDA Graph replay, speedups
are 4.14$\times$ on FastWAM and 7.15$\times$ on OpenWAM
(Table~\ref{tab:compiled-baselines}).

\textbf{Comparison with caching baselines.}
With compilation and CUDA Graph replay, TeaCache is the fastest competing
baseline at 49.28/245.99\,ms on
FastWAM/OpenWAM, with lower average success rates than ours on both.
BAC retains a higher OpenWAM success rate (87.77\% versus 87.41\%) at
305.08\,ms. Appendix~\ref{app:repo-ablation}
compares original and compiled execution for each baseline.

\subsection{Sources of latency reduction}
\label{sec:execution-ablation}

Table~\ref{tab:execution-stages} adds execution optimizations at fixed
evaluation budgets: first-forward timings come from 2F runs, and complete
requests from separate 10F runs. After graph replay and compilation,
parallel execution reduces first-forward latency from 36.21 to 17.77\,ms
on FastWAM and from 60.62 to 37.12\,ms on OpenWAM. Token reuse further
reduces these to 17.69 and 31.62\,ms; its incremental benefit is larger
on OpenWAM in this configuration.

Request latency with 10F falls from 214.33 to 29.47\,ms on FastWAM and
672.83 to 217.24\,ms on OpenWAM, establishing execution savings before
adaptive evaluation counts are introduced. Detailed stage and preparation
measurements appear in Appendix~\ref{app:rr-protocol}.

\subsection{Ablation studies}
\label{sec:prefill-results}\label{sec:motion-analysis}

Table~\ref{tab:component-ablation} ablates the full OpenWAM method.
Token ablations keep the motion gate unchanged; motion ablations keep
the token policy unchanged. Request latencies are averaged over complete
rollouts. Fixed 2F/4F use fixed evaluation counts with the normal token policy.

\begin{table}[!htbp]

\caption{\textbf{Component ablations and latency diagnostics.}
Left: OpenWAM/RoboTwin success rate, measured first-forward latency (First),
and rollout-mean request latency (Request); latencies are in ms.
Fixed 2F/4F retain the normal token policy.
$\checkmark$/$\times$: enabled/disabled; n/a: inactive;
Sel.: RGB/feature selection; Gate: motion criterion;
Ada./Rnd.: adaptive/random 2F/4F.
Right: (a) historical OpenWAM first-stage profile (line), with measured
first-forward latency for w/o filling (65 tokens) and Ours (80 tokens)
marked by diamonds; (b) rollout-mean request latencies for Fixed 2F,
Fixed 4F, and Ours, on RTX 4090.}
\label{tab:component-ablation}
\centering
\begin{minipage}[c]{0.70\linewidth}
\centering
\begingroup
\normalsize
\setlength{\tabcolsep}{0pt}
\renewcommand{\arraystretch}{1.0}
\newcommand{\ablon}{\textcolor[HTML]{52886F}{$\checkmark$}}
\newcommand{\abloff}{\textcolor[HTML]{BB655A}{$\times$}}
\newcommand{\ablna}{\textcolor[HTML]{89949A}{n/a}}
\small
\begin{tabular}{@{}p{0.275\linewidth}>{\centering\arraybackslash}p{0.095\linewidth}>{\centering\arraybackslash}p{0.08\linewidth}>{\centering\arraybackslash}p{0.06\linewidth}>{\centering\arraybackslash}p{0.08\linewidth}>{\centering\arraybackslash}p{0.08\linewidth}>{\centering\arraybackslash}p{0.1\linewidth}>{\centering\arraybackslash}p{0.11\linewidth}>{\centering\arraybackslash}p{0.12\linewidth}@{}}
\toprule
\multirow{2}{*}{Configuration} & \multicolumn{3}{c}{Token policy} & \multicolumn{2}{c}{Motion} & \multicolumn{3}{c}{Results} \\
\cmidrule(lr){2-4}\cmidrule(lr){5-6}\cmidrule(l){7-9}
& Reuse & Sel. & Fill & NFE & Gate & \mbox{SR (\%)} & First & Request \\
\midrule
\rowcolor{tablehighlight}
\textbf{Ours (full)} & \ablon & \ablon & \ablon & \textbf{Ada.} & \ablon & \textbf{93.4} & \textbf{30.96} & \textbf{63.76} \\
\midrule
w/o token reuse & \abloff & \ablna & \ablna & Ada. & \ablon & 93.8 & 37.41 & 72.10 \\
Random tokens & \ablon & \abloff & \ablon & Ada. & \ablon & 91.3 & 30.75 & 62.32 \\
w/o capacity filling & \ablon & \ablon & \abloff & Ada. & \ablon & 92.1 & 30.76 & 61.78 \\
\midrule
Random gate & \ablon & \ablon & \ablon & Rnd. & \abloff & 91.1 & 31.02 & 64.77 \\
Fixed 2F & \ablon & \ablon & \ablon & 2F & \abloff & 89.6 & 31.11 & 54.70 \\
Fixed 4F & \ablon & \ablon & \ablon & 4F & \abloff & 93.7 & 30.87 & 95.26 \\
\bottomrule
\end{tabular}
\endgroup
\end{minipage}\hfill
\begin{minipage}[c]{0.28\linewidth}
\centering
\includegraphics[width=\linewidth]{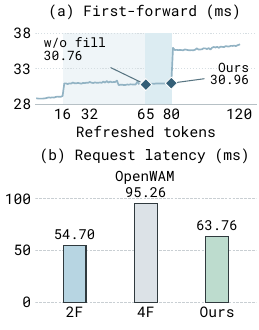}
\end{minipage}

\end{table}

\textbf{Token reuse and selection.}
Disabling reuse gives 93.8\% success at 72.10\,ms, compared with 93.4\%
at 63.76\,ms for the full method. Uniformly selecting $b_t$ random positions instead of the
RGB/feature-guided set preserves the refreshed token count but lowers the success rate to 91.3\%,
supporting informative token selection.

\textbf{Capacity filling.}
Filling from the minimum refresh budget $r_t$ to capacity $b_t$ raises
the success rate from 92.1\% to 93.4\%, with request latency increasing from 61.78
to 63.76\,ms. First-forward latency rises from 30.76 to 30.96\,ms
when filling from 65 to 80 tokens (Table~\ref{tab:component-ablation}a).

\textbf{Motion-adaptive refinement.}
The motion gate yields 93.4\% success at 63.76\,ms, versus
91.1\% at 64.77\,ms for random 2F/4F selection using the original gate's
usage proportions. This supports allocating refinement based on motion.
Fixed 2F reaches 89.6\% success at 54.70\,ms, while Fixed 4F reaches
93.7\% at 95.26\,ms. The adaptive policy attains a success rate close to Fixed 4F
with a lower mean request latency. Appendix~\ref{app:motion-motivation}
gives paired prediction diagnostics.

\textbf{Visualization of dynamic allocation.}
Figure~\ref{fig:dynamic-allocation} aligns observations, selected evaluation budgets,
and token allocation over one complete rollout. Both evaluation budgets
and all three token capacities occur.

\begin{figure}[!htbp]
\centering
\includegraphics[width=\textwidth]{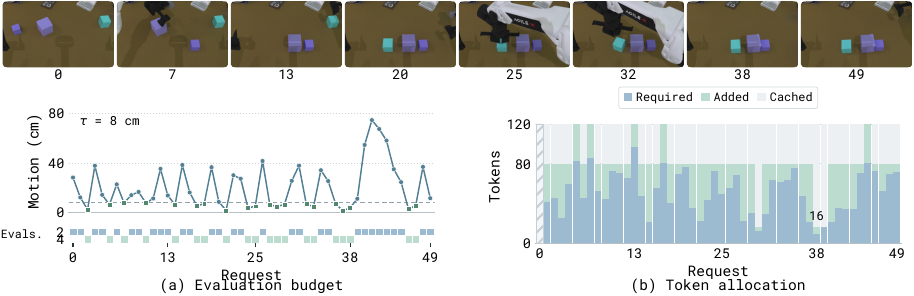}
\caption{\textbf{Dynamic allocation over one rollout.}
Shared observations accompany evaluation budgets (a) and token allocation (b)
across all 50 requests of an OpenWAM block-sorting rollout.
Hatching marks initialization.}
\label{fig:dynamic-allocation}
\end{figure}

\clearpage
\endgroup
\begin{figure}[!t]
\centering
\includegraphics[width=\textwidth]{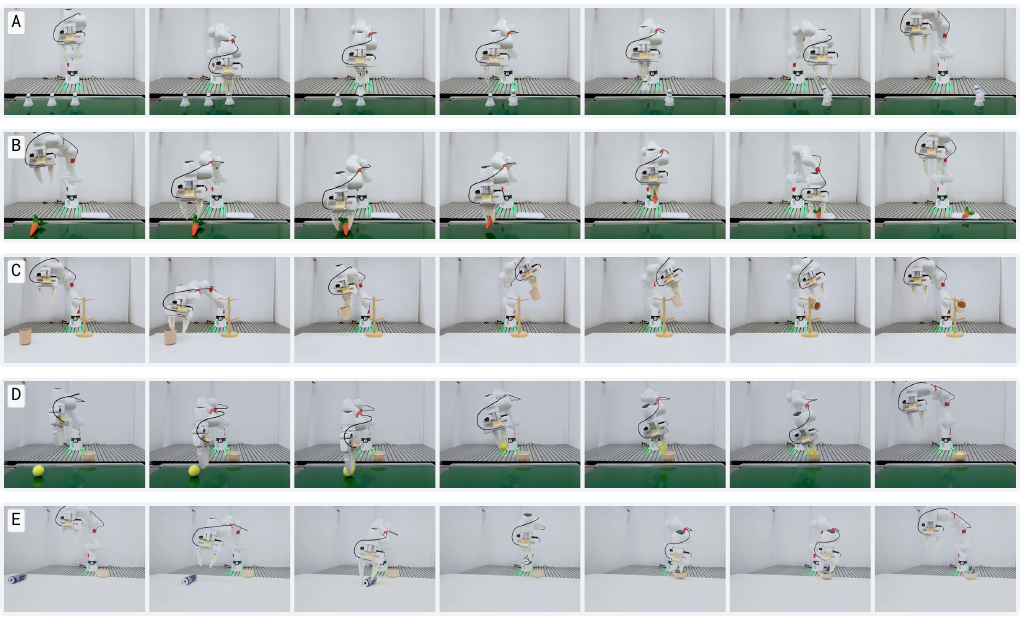}
\caption{\textbf{Real-world task sequences.}
Rows A--E show stacking three shuttlecocks, placing a carrot on a plate,
hanging a mug, transferring a tennis ball, and intercepting a rolling can,
respectively. Seven frames progress from left to right within each recording.}
\label{fig:real-world-sequences}
\end{figure}

\subsection{Real-world evaluation}
\label{sec:closed-loop}\label{sec:real-robot}
Table~\ref{tab:real-world-tasks} compares Native and \method on the five tasks
in Figure~\ref{fig:real-world-sequences} under the same synchronous
inference and action execution policy.
Average success rates rise from 35.6\% to 52.8\% on FastWAM and 33.6\% to
70.8\% on OpenWAM (+17.2 and +37.2 percentage points).
Mean inference latency decreases from 210.71 to 23.72\,ms and from
640.87 to 65.27\,ms, respectively.

\begin{table}[!ht]

\caption{\textbf{Real-world manipulation results.}
Success rate (SR, \%) and mean inference latency (Lat., ms).
Average weights tasks equally.}
\label{tab:real-world-tasks}
\begin{center}
\begingroup
\normalsize
\setlength{\tabcolsep}{0pt}
\renewcommand{\arraystretch}{1.05}
\small
\begin{tabular}{@{}p{0.24\linewidth}*{8}{>{\centering\arraybackslash}p{0.095\linewidth}}@{}}
\toprule
\multirow{3}{*}{Task} & \multicolumn{4}{c}{FastWAM} & \multicolumn{4}{c}{OpenWAM} \\
\cmidrule(lr){2-5}\cmidrule(l){6-9}
 & \multicolumn{2}{c}{Native} & \multicolumn{2}{c}{\textbf{Ours}} & \multicolumn{2}{c}{Native} & \multicolumn{2}{c}{\textbf{Ours}} \\
\cmidrule(lr){2-3}\cmidrule(lr){4-5}\cmidrule(lr){6-7}\cmidrule(l){8-9}
 & SR $\uparrow$ & Lat. $\downarrow$ & \textbf{SR} $\uparrow$ & \textbf{Lat.} $\downarrow$ & SR $\uparrow$ & Lat. $\downarrow$ & \textbf{SR} $\uparrow$ & \textbf{Lat.} $\downarrow$ \\
\midrule
Three shuttlecocks & 24.0 & 210.36 & \textbf{40.0} & \textbf{24.33} & 4.0 & 640.28 & \textbf{68.0} & \textbf{68.61} \\
Carrot to plate & 46.0 & 209.74 & \textbf{74.0} & \textbf{21.86} & 66.0 & 638.56 & \textbf{96.0} & \textbf{62.12} \\
Hang mug & 72.0 & 211.08 & \textbf{70.0} & \textbf{25.07} & 82.0 & 642.15 & \textbf{82.0} & \textbf{65.36} \\
Tennis ball to bowl & 26.0 & 210.52 & \textbf{34.0} & \textbf{23.24} & 16.0 & 639.47 & \textbf{76.0} & \textbf{64.85} \\
Rolling can to bowl & 10.0 & 211.85 & \textbf{46.0} & \textbf{24.11} & 0.0 & 643.89 & \textbf{32.0} & \textbf{65.43} \\
\midrule
Average & 35.6 & 210.71 & \textbf{52.8} & \textbf{23.72} & 33.6 & 640.87 & \textbf{70.8} & \textbf{65.27} \\
\bottomrule
\end{tabular}
\endgroup
\end{center}
\end{table}

\section{Conclusion}

\method combines parallel execution, hardware-aware token reuse,
and motion-adaptive refinement in a training-free framework.
On FastWAM and OpenWAM, measured average speedups reach 8.90$\times$
and 10.67$\times$, with success rates within 0.02 and 0.53 percentage
points of native inference. Across five real-world tasks, average success rates
improve by 17.2 and 37.2 percentage points.
\par\label{page:main-text-end}

\clearpage
\endgroup
\subsection*{AI use statement}
We used generative AI tools to assist with manuscript writing and editing, including improvements
to wording, grammar, clarity, and organization, as well as LaTeX formatting. The authors reviewed
and revised the AI-assisted content to ensure that it accurately reflects the methods and results. The
authors take full responsibility for the accuracy and integrity of the final manuscript.

\bibliographystyle{iclr2027_conference}
\bibliography{references}
\clearpage
\appendix
\section{Real-World Setup and Evaluation}
\label{app:real-world-setup}

\subsection{Platforms and tasks}

\begin{figure}[!ht]
\centering
\includegraphics[width=\textwidth]{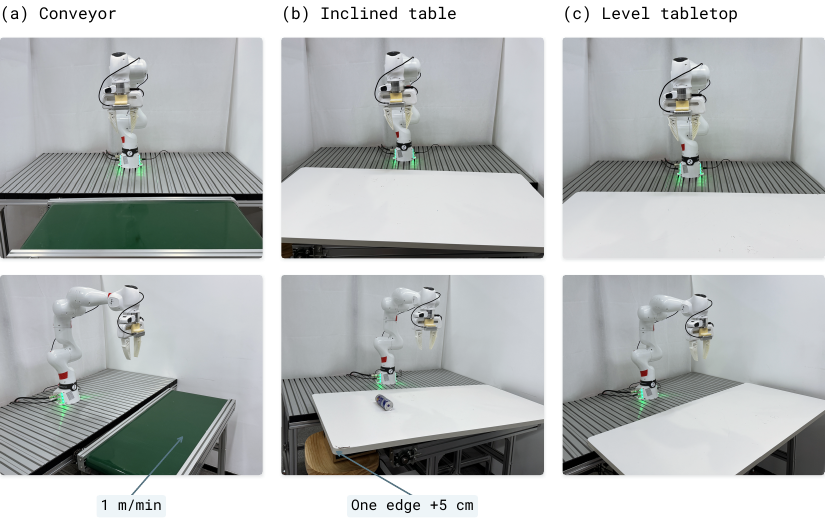}
\caption{\textbf{Real-world experimental setups.}
Front views (top) and oblique views (bottom) of the three setups.
(a) Conveyor. (b) Inclined tabletop. (c) Level tabletop.}
\label{fig:real-world-setup}
\end{figure}

We use a Franka Research 3 robot and a RealSense D435i camera.
Figure~\ref{fig:real-world-setup} shows the three setups used for our five
tasks. The conveyor runs at 1\,m/min for shuttlecock stacking, carrot
placement, and tennis-ball placement. The rolling-can task uses an
inclined tabletop with one edge raised by 5\,cm; mug hanging uses a level
tabletop.

\subsection{Evaluation protocol}

We compare native inference with \method on both FastWAM and OpenWAM.
Each task uses 50 trials per backbone and inference configuration, with the
same synchronous inference and action execution policy across the comparison.
Each inference request predicts a chunk of 32 future actions. The robot
executes actions at a control frequency of 20\,Hz. Inference runs locally,
with negligible network transmission latency.
Table~\ref{tab:real-world-protocol} lists the task goals; its task IDs match
the rollout rows in Figure~\ref{fig:real-world-sequences}.
Table~\ref{tab:real-world-tasks} reports success rate and measured mean
inference latency; its averages weight the five tasks equally.

In our conveyor experiments, task success was substantially lower when
measured inference latency exceeded approximately 90\,ms. The dashed line
in Figure~\ref{fig:opening} marks this empirically observed latency threshold.

\begin{table}[!ht]
\caption{\textbf{Real-world task goals and trial counts.}
Trials are counted per task, backbone, and inference configuration.}
\label{tab:real-world-protocol}
\begin{center}
\small
\setlength{\tabcolsep}{3pt}
\renewcommand{\arraystretch}{1.2}
\begin{tabular}{@{}p{87pt}p{53pt}p{194pt}>{\centering\arraybackslash}p{44pt}@{}}
\toprule
Task & Setup & Task goal & Trials \\
\midrule
A: Three shuttlecocks & Conveyor & Stack the three shuttlecocks together. & 50 \\
B: Carrot to plate & Conveyor & Move the carrot from the belt onto the plate. & 50 \\
C: Hang mug & Level tabletop & Hang the mug by its handle on the rack. & 50 \\
D: Tennis ball to bowl & Conveyor & Move the tennis ball from the belt into the bowl. & 50 \\
E: Rolling can to bowl & Inclined table & Intercept the rolling can and place it in the bowl. & 50 \\
\bottomrule
\end{tabular}
\end{center}
\end{table}

\section{Evaluation Protocols and Detailed Results}
\label{app:evaluation-slots}

\subsection{Benchmarks and comparison settings}

\paragraph{Benchmarks.}
We evaluate FastWAM and OpenWAM on RoboTwin 2.0~\citep{chen2025robotwin2},
LIBERO~\citep{liu2023libero}, and LIBERO-Plus~\citep{fei2026liberoplus}.
RoboTwin Clean and Random are reported separately, giving four settings
in Table~\ref{tab:main-comparison}. LIBERO and LIBERO-Plus each report
Spatial, Object, Goal, and Long categories; their individual success rates
appear in Table~\ref{tab:benchmark-details}.

\paragraph{Evaluation protocol.}
We use the benchmark environments' task-success checks. A rollout succeeds
if the task goal is satisfied within the evaluation horizon; otherwise it
fails. RoboTwin uses 100 rollouts per task in each of its Clean and Random
settings. LIBERO uses 50 rollouts per task with the released initial-state
sets and goal predicates; Long denotes \texttt{libero\_10}. LIBERO-Plus
uses one rollout per perturbed task variant, covering changes to layout,
cameras, robot initial state, language, lighting, background, and sensor
noise~\citep{fei2026liberoplus}. These protocols apply to both backbones
and all methods in Table~\ref{tab:main-comparison}.

\paragraph{Comparison settings.}
We use the officially released fine-tuned checkpoints of FastWAM and
OpenWAM. For each backbone and evaluation setting, Native, the caching
baselines, and \method share the same model weights.
All methods use the same evaluation seeds and initial scenes within each
backbone and evaluation setting, enabling paired comparisons.
Each backbone's original Native execution provides the reference.
VLA-Cache, TeaCache, ToCa, BAC, and \method use compilation and CUDA Graph
replay for the main-table latency comparison. The caching baselines retain
their original execution schedules, while \method incorporates the proposed
parallel execution. Appendix~\ref{app:repo-ablation}
compares original and compiled execution, including Native.
Runtime measurements use BF16, batch size one, and a modified
48\,GiB RTX 4090.

\subsection{Metrics and aggregation}

For an evaluated task or task variant $q$, let $n_q$ be the number of
rollouts and $y_{q,i}\in\{0,1\}$ the benchmark's success indicator.
Its success rate is $\mathrm{SR}_q=100\,n_q^{-1}\sum_{i=1}^{n_q}y_{q,i}$.
RoboTwin averages task success rates within each setting. LIBERO and
LIBERO-Plus average task or task-variant success rates within each category;
their main-table scores are the arithmetic means of the four categories.
For each setting, latency is the mean of measured request times
collected during complete closed-loop rollouts. Each request uses its
selected token set and Transformer evaluation budget.
The execution comparisons in Tables~\ref{tab:execution-stages},
\ref{tab:rr-full-matrix}, and~\ref{tab:compiled-baselines} use the same
rollout-based mean latency definition for each evaluated configuration.

Timing runs from prepared CPU images and robot state to denormalized CPU
actions. It includes input processing, model and sampler execution,
transfers, and synchronization. Compilation, graph capture, warmup,
profiling, and robot execution are excluded.

Tables~\ref{tab:main-comparison} and~\ref{tab:compiled-baselines} average
latency equally across the four settings; Table~\ref{tab:main-comparison}
also averages success equally across these settings.
Speedup is the original Native average latency divided
by the method's average latency, computed before rounding.
Table~\ref{tab:component-ablation} reports measured first-forward latency
(First) and mean request latency over complete rollouts (Request).
All latencies are reported in milliseconds.

\subsection{Component ablation settings}

Table~\ref{tab:component-ablation} evaluates variants of the full OpenWAM
method on RoboTwin. Token ablations keep the motion gate unchanged; motion
ablations keep the token policy unchanged.

\paragraph{Token reuse and selection.}
\emph{w/o token reuse} refreshes all observation positions.
\emph{Random tokens} retains the selected execution capacity $b_t$ and uniformly samples
that many positions without replacement from all $M$ positions.
\emph{w/o capacity filling} refreshes $r_t$ positions instead of $b_t$,
retaining required positions and feature ranking.

\paragraph{Motion gate.}
\emph{Fixed 2F} and \emph{Fixed 4F} use two and four Transformer evaluations,
respectively, within the ten Euler updates. Their token refresh counts
follow the observation. \emph{Random gate} samples between the two budgets
using the original motion gate's usage proportions, independently of
predicted motion.

\subsection{Category-level benchmark results}
\label{app:benchmark-details}
Table~\ref{tab:benchmark-details} gives the category-level success rates
underlying the benchmark summaries in Table~\ref{tab:main-comparison}.
The RoboTwin average combines Clean and Random; each LIBERO or Plus
average combines Spatial, Object, Goal, and Long with equal weight.
\begin{table}[!ht]

\caption{\textbf{Category-level success rates (\%).} Each backbone uses the same
methods and success-rate evaluations as Table~\ref{tab:main-comparison}.
Plus denotes LIBERO-Plus.}
\label{tab:benchmark-details}
\begin{center}
\normalsize
\setlength{\tabcolsep}{1pt}
\renewcommand{\arraystretch}{1.08}
\begingroup
\small
\begin{tabular}{@{}p{52pt}p{35pt}>{\centering\arraybackslash}p{50pt}*{4}{>{\centering\arraybackslash}p{\dimexpr(\linewidth-210pt)/4\relax}}>{\columncolor{tablehighlight}\centering\arraybackslash}p{59pt}@{}}
\multicolumn{8}{@{}l}{\textbf{(a) FastWAM}} \\
\midrule
Benchmark & Subset & Native & VLA-Cache & TeaCache & ToCa & BAC & \textbf{Ours} \\
\midrule
\multirow{3}{*}{RoboTwin} & Clean & \mbox{91.88} & \mbox{80.00} & \mbox{81.42} & \mbox{91.35} & \mbox{91.67} & \textbf{92.04} \\
 & Random & \mbox{91.78} & \mbox{83.20} & \mbox{90.86} & \mbox{91.24} & \mbox{89.53} & \textbf{90.50} \\
 & Avg. SR & \mbox{91.83} & \mbox{81.60} & \mbox{86.14} & \mbox{91.30} & \mbox{90.60} & \textbf{91.27} \\
\cmidrule(lr){1-8}
\multirow{5}{*}{LIBERO} & Spatial & \mbox{98.20} & \mbox{98.00} & \mbox{97.80} & \mbox{94.40} & \mbox{98.20} & \textbf{96.20} \\
 & Object & \mbox{100.00} & \mbox{100.00} & \mbox{99.80} & \mbox{98.80} & \mbox{100.00} & \textbf{99.60} \\
 & Goal & \mbox{97.00} & \mbox{96.00} & \mbox{96.20} & \mbox{93.20} & \mbox{96.80} & \textbf{97.60} \\
 & Long & \mbox{95.20} & \mbox{94.00} & \mbox{94.40} & \mbox{91.80} & \mbox{95.00} & \textbf{95.60} \\
 & Avg. SR & \mbox{97.60} & \mbox{97.00} & \mbox{97.05} & \mbox{94.55} & \mbox{97.50} & \textbf{97.25} \\
\cmidrule(lr){1-8}
\multirow{5}{*}{Plus} & Spatial & \mbox{52.00} & \mbox{52.80} & \mbox{53.20} & \mbox{53.00} & \mbox{54.60} & \textbf{53.60} \\
 & Object & \mbox{69.80} & \mbox{68.80} & \mbox{69.20} & \mbox{69.20} & \mbox{71.00} & \textbf{71.00} \\
 & Goal & \mbox{37.00} & \mbox{35.60} & \mbox{36.40} & \mbox{36.40} & \mbox{38.80} & \textbf{38.00} \\
 & Long & \mbox{40.40} & \mbox{37.20} & \mbox{38.00} & \mbox{38.00} & \mbox{41.80} & \textbf{42.20} \\
 & Avg. SR & \mbox{49.80} & \mbox{48.60} & \mbox{49.20} & \mbox{49.15} & \mbox{51.55} & \textbf{51.20} \\
\bottomrule
\end{tabular}
\endgroup
\par\vspace{10pt}
\begingroup
\small
\begin{tabular}{@{}p{52pt}p{35pt}>{\centering\arraybackslash}p{50pt}*{4}{>{\centering\arraybackslash}p{\dimexpr(\linewidth-210pt)/4\relax}}>{\columncolor{tablehighlight}\centering\arraybackslash}p{59pt}@{}}
\multicolumn{8}{@{}l}{\textbf{(b) OpenWAM}} \\
\midrule
Benchmark & Subset & Native & VLA-Cache & TeaCache & ToCa & BAC & \textbf{Ours} \\
\midrule
\multirow{3}{*}{RoboTwin} & Clean & \mbox{93.64} & \mbox{93.36} & \mbox{90.42} & \mbox{93.42} & \mbox{93.52} & \textbf{92.78} \\
 & Random & \mbox{94.06} & \mbox{89.94} & \mbox{94.12} & \mbox{93.98} & \mbox{91.96} & \textbf{94.02} \\
 & Avg. SR & \mbox{93.85} & \mbox{91.65} & \mbox{92.27} & \mbox{93.70} & \mbox{92.74} & \textbf{93.40} \\
\cmidrule(lr){1-8}
\multirow{5}{*}{LIBERO} & Spatial & \mbox{99.60} & \mbox{98.80} & \mbox{99.00} & \mbox{98.40} & \mbox{99.20} & \textbf{99.40} \\
 & Object & \mbox{99.60} & \mbox{99.40} & \mbox{99.40} & \mbox{99.20} & \mbox{99.60} & \textbf{99.60} \\
 & Goal & \mbox{99.80} & \mbox{98.60} & \mbox{98.80} & \mbox{98.20} & \mbox{99.00} & \textbf{99.20} \\
 & Long & \mbox{98.20} & \mbox{96.80} & \mbox{97.20} & \mbox{96.60} & \mbox{97.80} & \textbf{97.80} \\
 & Avg. SR & \mbox{99.30} & \mbox{98.40} & \mbox{98.60} & \mbox{98.10} & \mbox{98.90} & \textbf{99.00} \\
\cmidrule(lr){1-8}
\multirow{5}{*}{Plus} & Spatial & \mbox{80.80} & \mbox{76.80} & \mbox{80.80} & \mbox{80.80} & \mbox{84.80} & \textbf{81.40} \\
 & Object & \mbox{64.00} & \mbox{59.80} & \mbox{64.00} & \mbox{64.00} & \mbox{60.00} & \textbf{65.80} \\
 & Goal & \mbox{42.20} & \mbox{42.20} & \mbox{46.40} & \mbox{46.20} & \mbox{54.00} & \textbf{39.60} \\
 & Long & \mbox{72.00} & \mbox{72.00} & \mbox{64.00} & \mbox{63.80} & \mbox{68.00} & \textbf{68.60} \\
 & Avg. SR & \mbox{64.75} & \mbox{62.70} & \mbox{63.80} & \mbox{63.70} & \mbox{66.70} & \textbf{63.85} \\
\bottomrule
\end{tabular}
\endgroup
\end{center}
\end{table}

\clearpage
\section{Execution Efficiency}
\label{app:rr-protocol}\label{app:rr-numerics}\label{app:rr-ledger}\label{app:rr-timeline}

\subsection{Execution settings and fixed budgets}
Table~\ref{tab:execution-stages} cumulatively adds CUDA Graph replay,
compilation, parallel execution, and token reuse. First starts from CPU images
and state and ends after input processing, Prefill, and the first complete
predictive evaluation. It includes selection, gathering, cache writes,
and stream joins. First is measured in 2F runs; Full 10F uses separate
complete requests through denormalized CPU action return. Both report
mean wall-clock latency measured during complete closed-loop rollouts,
excluding compilation, capture, warmup, and profiling.

Table~\ref{tab:rr-full-matrix} compares dense execution at two, four, and
ten Transformer evaluations within ten Euler updates. Measurements use
BF16 on the modified 48\,GiB RTX 4090, PyTorch 2.7.1/CUDA 12.8, an Intel
Core Ultra 9 285K, four PyTorch threads, and CPU affinity 0--7.
OpenWAM's Native baseline uses commit \texttt{a8992b613fdb}; compilation
uses \texttt{reduce-overhead} with verified graph replay. Parallel denotes
dependency-aware scheduling and its accompanying runtime optimizations
(Section~\ref{sec:parallel-execution}).

OpenWAM timing begins with composed CPU PIL images and state;
FastWAM begins with three RGB arrays and state, including composition
and resize. Both include input preparation, VAE and context processing,
model evaluations, sampler updates, transfers, synchronization, and CPU
action return. Prompt embeddings are warm. Sensor acquisition,
communication, robot execution, setup, and output-video decoding are excluded.

\begin{table}[!ht]
\normalsize
\caption{\textbf{Dense execution controls at fixed evaluation budgets.}
Mean request latency measured during complete closed-loop rollouts
in ms, following the main-text timing protocol.
All three execution variants compute full FFNs.}
\label{tab:rr-full-matrix}\label{tab:runtime-foundation}
\centering
\small
\begin{tabular*}{\linewidth}{@{\extracolsep{\fill}}lrlll@{}}
\toprule
Model & $N$ & Native & Compile + Graph & Parallel \\
\midrule
\rrFullMatrixRows
\bottomrule
\end{tabular*}
\end{table}

At ten evaluations, the Parallel configuration reduces Compile + Graph latency from
450.90 to 223.27\,ms on OpenWAM and from 99.68 to 33.52\,ms on FastWAM.

\subsection{Numerical agreement}
Table~\ref{tab:rr-numerics} compares Parallel and Native predictions
at matched budgets. FastWAM checks vary state, instruction, images, and
noise; direct execution and graph replay agree bitwise on these cases.
OpenWAM's official and modified Native controls also agree bitwise.

\begin{table}[!ht]
\normalsize
\caption{\textbf{Parallel versus native output checks at matched budgets.}
Mean case MAE averages recorded action MAEs; Max is the largest coordinate
difference. Each model uses its own effective action space.}
\label{tab:rr-numerics}
\centering
\small
\begin{tabular*}{\linewidth}{@{\extracolsep{\fill}}lrrrr@{}}
\toprule
Model & $N$ & Cases & Mean case MAE & Max \\
\midrule
\rrNumericalRows
\bottomrule
\end{tabular*}
\end{table}
\clearpage
\subsection{Compiled caching baselines}
\label{app:repo-ablation}
Table~\ref{tab:compiled-baselines} compares original execution with
compilation and CUDA Graph replay under the main-table evaluation protocol.
Native remains the original implementation when computing main-table speedup.

\begin{table}[!ht]
\caption{\textbf{Effect of compilation and CUDA Graph replay on inference latency.}
Original denotes each method's original implementation; + C/G enables
compilation and CUDA Graph replay with its execution schedule preserved.
All methods report measured
rollout-mean latencies (ms), with task settings, timing boundaries, and
four-setting averaging following Table~\ref{tab:main-comparison}.
Native uses the 10F results in Table~\ref{tab:rr-full-matrix}.
Reduction is the percentage decrease
from Original to + C/G.}
\label{tab:compiled-baselines}
\begin{center}
\small
\setlength{\tabcolsep}{0pt}
\renewcommand{\arraystretch}{1.08}
\begin{tabular}{@{}p{0.22\linewidth}*{6}{>{\centering\arraybackslash}p{0.13\linewidth}}@{}}
\toprule
\multirow{2}{*}{Method} & \multicolumn{3}{c}{FastWAM} & \multicolumn{3}{c}{OpenWAM} \\
\cmidrule(lr){2-4}\cmidrule(l){5-7}
 & Original & + C/G & Reduction & Original & + C/G & Reduction \\
\midrule
Native & 214.33 & 99.68 & 53.5\% & 672.83 & 450.90 & 33.0\% \\
\midrule
VLA-Cache & 236.13 & 115.55 & 51.1\% & 743.32 & 400.26 & 46.2\% \\
TeaCache & \rankbest{112.85} & \rankbest{49.28} & 56.3\% & \rankbest{362.31} & \rankbest{245.99} & 32.1\% \\
ToCa & 225.50 & 102.66 & 54.5\% & 719.60 & 440.42 & 38.8\% \\
BAC & \rankrunnerup{151.17} & \rankrunnerup{67.14} & 55.6\% & \rankrunnerup{460.04} & \rankrunnerup{305.08} & 33.7\% \\
\bottomrule
\end{tabular}
\end{center}
\end{table}

\subsection{Preparation scheduling}
The historical OpenWAM step-600 profiling runtime uses BF16 on the
48\,GiB RTX 4090 with PyTorch 2.7.1/CUDA 12.8 (snapshot
\texttt{5f9dede}, upstream \texttt{898e2f9}). It refreshes 80 observation
positions, computes all future-video and action positions, and uses two
Transformer evaluations within ten Euler updates. Repeated inputs and
$\alpha=0$, $\mathcal B=\{80,120\}$ fix the refresh count; online selection
and filling still execute.
Two observations are each repeated for 50 rounds with three interleaved
schedules, restoring dense history and noise before every call. Timing
covers composed CPU images/state through CPU actions, including preparation,
selection, cache updates, transfers, and waits; restoration and warmup are excluded.

\begin{table}[!ht]
\normalsize
\caption{\textbf{Preparation scheduling at a fixed 80-token budget.}
Historical OpenWAM profiling, $N=2$. Each row summarizes 100 requests in ms.
First and later are GPU-event medians; request percentiles use wall time.}
\label{tab:repo-runtime}
\begin{center}
\small
\begin{tabular*}{\linewidth}{@{\extracolsep{\fill}}lrrrrr@{}}
\toprule
Preparation schedule & First & Later & Request P50 & P90 & P95 \\
\midrule
Sequential & 23.06 & 20.06 & 51.32 & 51.65 & 51.79 \\
VAE/condition overlap & 23.06 & 20.08 & 50.55 & 50.91 & 50.97 \\
Overlap + early RGB & 22.97 & 20.05 & 50.26 & 50.62 & 50.73 \\
\bottomrule
\end{tabular*}
\end{center}
\end{table}

The schedules select identical positions and return bitwise-identical actions.
Paired median savings are 0.778\,ms for VAE/condition overlap and
0.274\,ms for early RGB routing, with round-bootstrap 95\% intervals
[0.742, 0.822] and [0.225, 0.324]\,ms.

\clearpage
\section{Token Reuse and Selection}
\label{app:prefill-contract}\label{app:ffn-contract}\label{app:materials-selectors}
\label{app:token-selection-ablation}
\newcommand{\RefreshSweepCounts}{120}
\newcommand{\RefreshSweepRequests}{3,585}
\newcommand{\RefreshSweepBatches}{4}
\newcommand{\SweepFullFirst}{36.41}
\newcommand{\SweepFilledFirst}{30.95}
\newcommand{\SweepFullRequest}{55.94}
\newcommand{\SweepFilledRequest}{50.36}
\newcommand{\SweepBaseRequest}{50.71}
\newcommand{\SweepBaseMAE}{0.00503}
\newcommand{\SweepFilledMAE}{0.00389}
\newcommand{\SweepFillReduction}{22.7}
\newcommand{\SweepRequestGain}{10.0}
\newcommand{\SweepDenseFloor}{0.00113}

\subsection{Cache state and feature correspondence}
The observation cache stores ungated FFN outputs for 120 positions at
each layer. Layer 0 is dense; layers 1--29 share a refresh set. Attention
uses current features at every position, and current residual gates apply
to fresh and cached FFN outputs. Future-video and action branches are
complete at every requested evaluation. Selected cache entries update;
others retain their last refreshed outputs. The layer-0 selector reference
advances at each observation. Episode boundaries reset history; changes
to weights, conditioning, camera layout, or preprocessing require compatible
cache state or full refresh.

\begin{figure}[!ht]
\centering
\includegraphics[width=.91\textwidth]{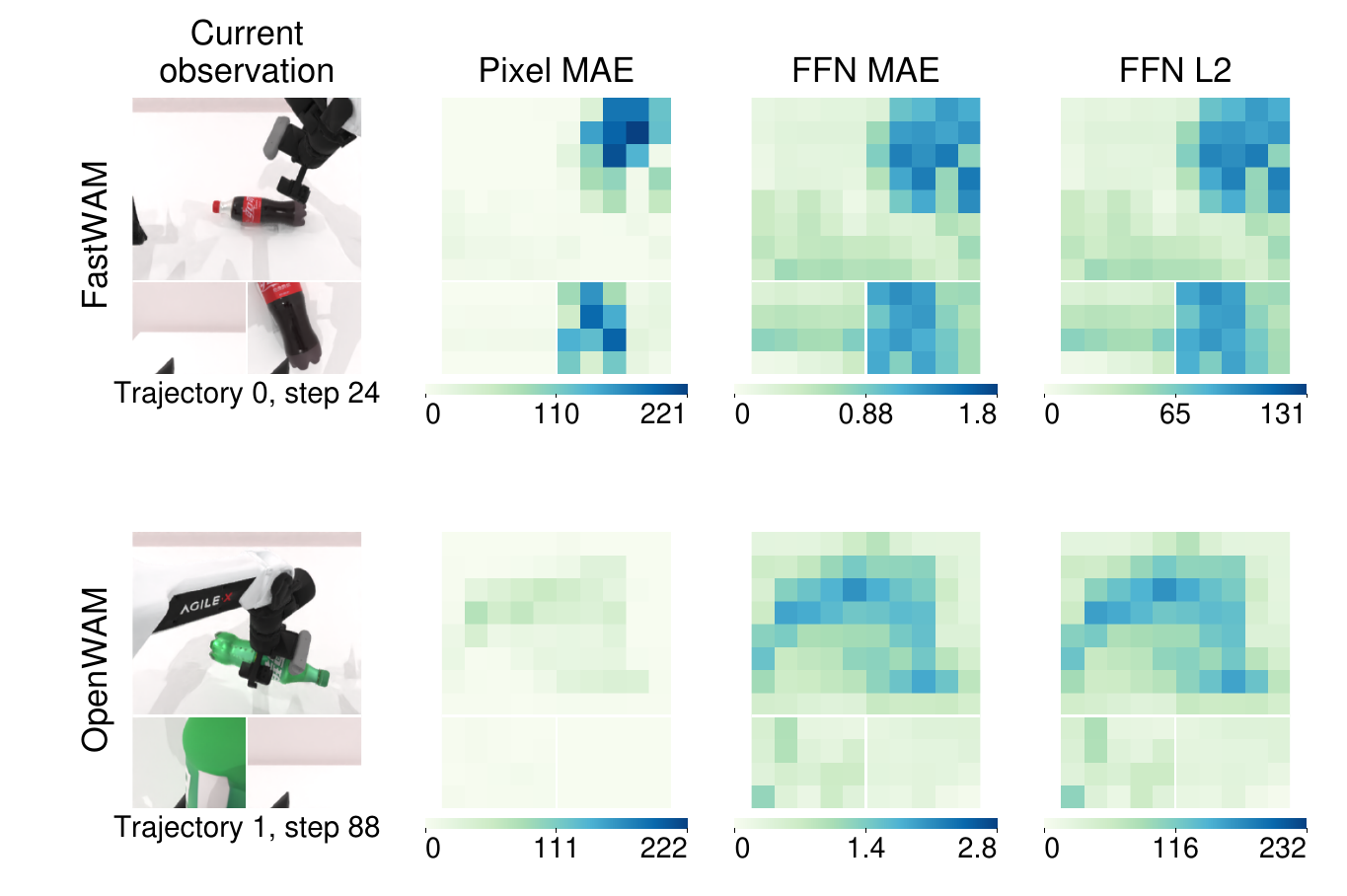}
\caption{\textbf{RGB and layer-0 FFN changes: typical examples.}
Observation pairs are eight environment steps apart. Camera regions are
spatially aligned; each model and metric uses a fixed color range.}
\label{fig:prefill-dense-first}
\end{figure}

\begin{figure}[!ht]
\centering
\includegraphics[width=.91\textwidth]{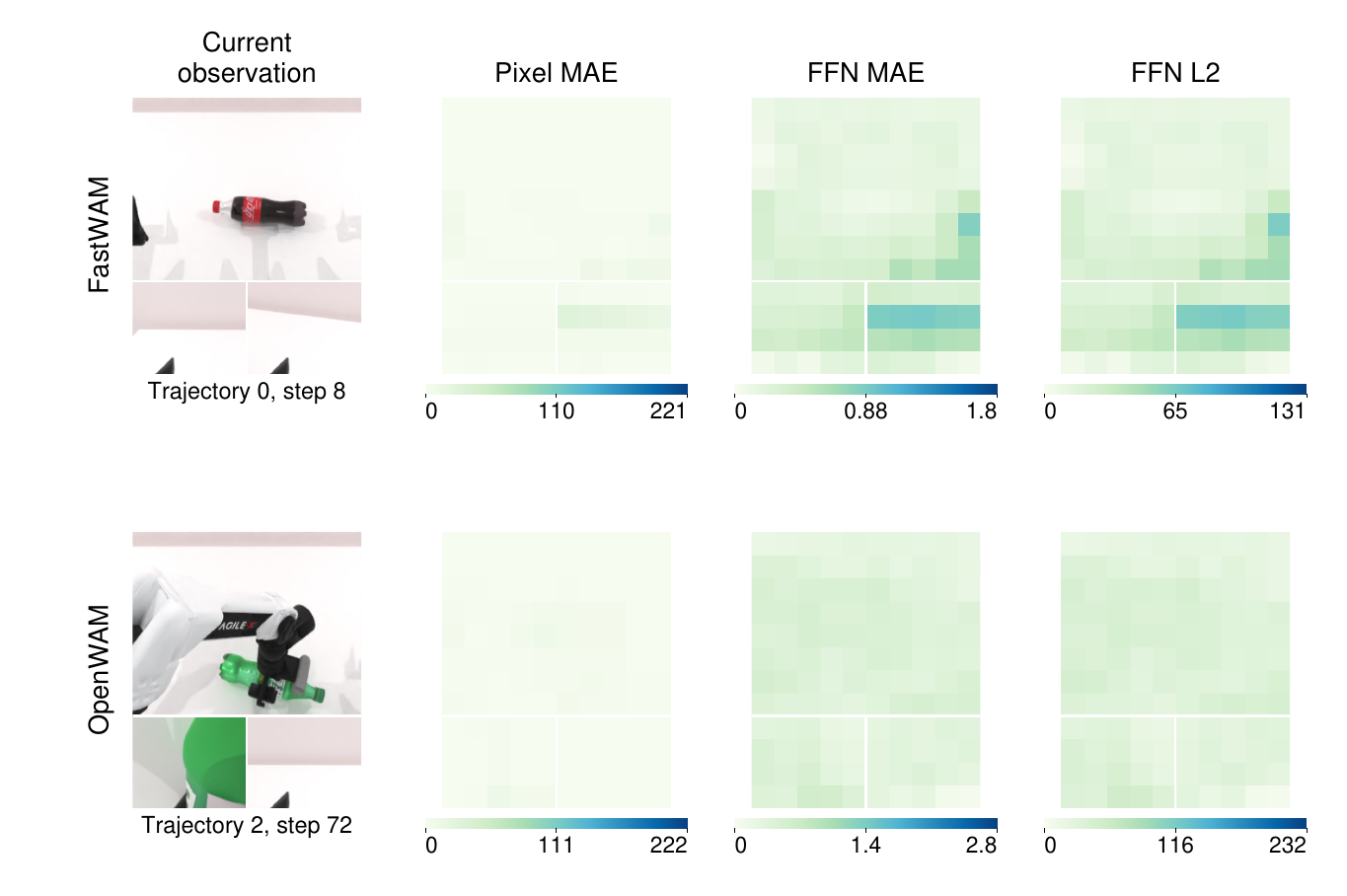}
\caption{\textbf{Examples with weaker spatial correspondence.}
Color ranges match Figure~\ref{fig:prefill-dense-first}; low pixel change
does not always imply low feature change.}
\label{fig:prefill-first-weak}
\end{figure}

We compare 14 observation pairs from one FastWAM trajectory and 44 pairs
from four OpenWAM \texttt{adjust\_bottle} trajectories, eight environment
steps apart. FFN change at each token is the channelwise mean absolute
or L2 difference between dense features. Mean within-frame RGB/FFN-MAE
correlations are 0.827/0.817 for FastWAM/OpenWAM at layer 0 and
0.716/0.701 across layers 1--29. Figures~\ref{fig:prefill-dense-first}
and~\ref{fig:prefill-first-weak} show typical and weaker correspondence:
small RGB changes can accompany larger feature changes. Each model and
metric uses the same color range across examples.

For the 44 OpenWAM pairs, a dense-history probe selects the 56 positions
with the smallest layer-0 change and compares them with later layers.
Change is measured by symmetric relative L1 distance,
$2\lVert x-y\rVert_1/\max(\lVert x\rVert_1+\lVert y\rVert_1,10^{-12})$.
The mean observation-set overlap is 85.4\%, versus 46.7\% by chance,
supporting a shared selection across observation layers.

\begin{figure}[!htb]
\centering
\includegraphics[width=\linewidth]{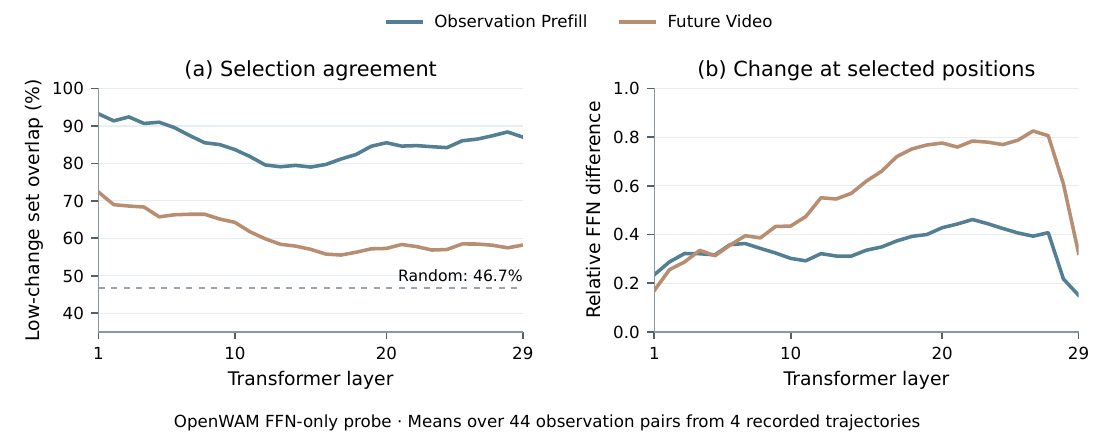}
\caption{\textbf{Stability of a selection across layers.}
(a) Overlap with each layer's lowest-change set. (b) Relative FFN change
at selected positions. Curves average 44 observation pairs. The future-video branch
is a separate diagnostic selection; its mean overlap is 61.0\%.}
\label{fig:appendix-layerwise-stability}
\end{figure}

\clearpage
\clearpage

\subsection{Selection at matched refresh counts}
\label{app:spatial-mask-replay}
\begin{figure}[!htbp]
\centering
\includegraphics[width=\textwidth]{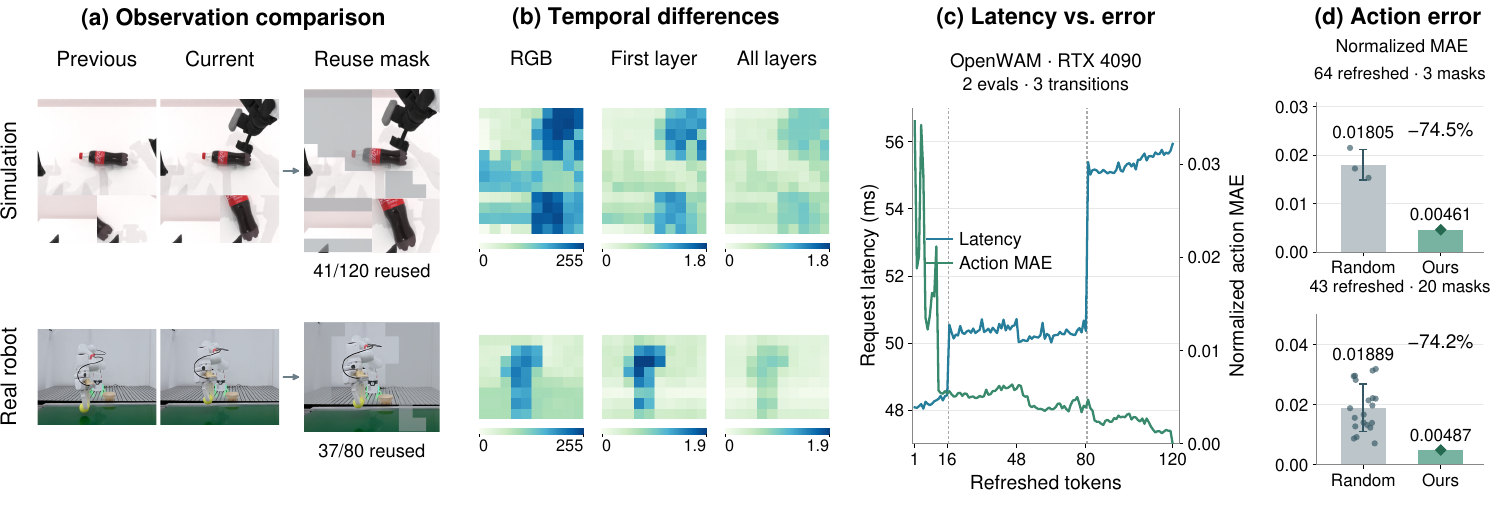}
\caption{\textbf{Token selection and reuse diagnostics.}
\textbf{(a,b)} Simulation and real-video observations, reuse masks, and
RGB/feature changes. \textbf{(c)} Historical OpenWAM profiling: request latency and normalized
action MAE versus refreshed-token count at $N=2$.
\textbf{(d)} Random versus RGB-guided selection at matched counts; bars
show mean action MAE, dots show individual masks, and whiskers show one
standard deviation.}
\label{fig:prefill-error}
\end{figure}

Figure~\ref{fig:prefill-error} compares RGB-guided and random selection
on one simulation pair and one real-video pair. Each comparison fixes
refresh count, cache history, inputs, noise, and the two Transformer
evaluations at updates 0 and 3. Layer 0 and attention remain complete.
Action MAE is measured against the same-input dense prediction over the
32-action chunk, using the model's active normalized coordinates.

\paragraph{Real video.}
The tennis-ball pair is sampled at 14.6 and 15.0\,s from a 120-fps
recording, selected by RGB MAE among candidate pairs 0.4\,s apart.
An $1800\times1440$ crop is resized to $320\times256$ in the front-camera
slot, with other camera rows fixed black. The step-3,500 OpenWAM checkpoint
is fine-tuned on the five robot tasks. Both frames use fixed proprioception
from frame 294 of a recorded tennis-ball trajectory, training normalization,
and the instruction ``Pick up the tennis ball from the conveyor and place
it into the bowl.'' BF16 feature capture and replay use A100.
Guided and 20 random masks each reuse 37 of 80 visible positions plus
40 padded positions. Mean action MAE over ten active coordinates is
0.004866 for guided selection and 0.018892 for random selection (74.2\% lower).

\paragraph{Simulation.}
The released three-camera FastWAM RoboTwin checkpoint uses transition
\texttt{fast\_ep0\_obs3} in A100 eager mode. Guided and three random masks
each refresh 64 of 120 positions. MAE over 14 normalized joint-action
coordinates is 0.004609 for guided selection and 0.018048 for random masks
on average (74.5\% lower). Full refresh matches dense computation bitwise.

\paragraph{Spatial masks.}
Panels a,b use an RGB-only diagnostic rule on the $12\times10$ grid:
refresh positions above $5/255$ mean absolute RGB change, then protect
the top quarter of remaining positions by a camera-local smoothed score.
Smoothing uses fourfold downsampling and a Gaussian with standard deviation
four and reflected boundaries. These displayed masks reuse 41/120 simulation
positions and 37/80 visible real-camera positions. Panel d uses the fixed
counts above. The full selector is evaluated through Random tokens in
Table~\ref{tab:component-ablation}.

\clearpage
\subsection{Capacity filling and recursive reuse}
\paragraph{Refresh-count sweep.}
Figure~\ref{fig:prefill-error}c uses the historical step-600 OpenWAM profiling
runtime on RTX 4090 with two Transformer evaluations. Three observation
pairs (frames 1598/1600, 2798/2800, and 3998/4000 of a 30-FPS recording)
are evaluated at all 120 refresh counts with five timing rounds per pair.
Each call restores dense history and noise. RGB-required positions precede
other positions, ranked within each group by layer-0 feature change.
The three pairs have 1, 0, and 12 required positions; smaller budgets
omit required positions in this fixed-budget diagnostic.

The \RefreshSweepRequests{} timed calls interleave target counts with an
80-token control. Each point uses the control median plus the median paired
latency difference. Timing includes online selection, packing, cache writes,
transfers, and waits; restoration, warmup, and error analysis are excluded.
Action MAE averages the $32\times10$ normalized action coordinates over
three pairs relative to full refresh. That reference has mean MAE
\SweepDenseFloor{} against the original dense engine.

Reducing refresh from 120 to 80 changes First from \SweepFullFirst{} to
\SweepFilledFirst{}\,ms and Request from \SweepFullRequest{} to
\SweepFilledRequest{}\,ms. Within the middle plateau, 32-to-80 filling
reduces MAE from \SweepBaseMAE{} to \SweepFilledMAE{} (\SweepFillReduction{}\%),
with request latencies \SweepBaseRequest{} and \SweepFilledRequest{}\,ms;
individual error increments are not monotonic. Table~\ref{tab:component-ablation}a
separately marks measured First at 65 and 80 tokens (30.76/30.96\,ms).
Its Request column contains complete-rollout means (61.78/63.76\,ms).

\paragraph{Recursive reuse.}
\label{app:prediction-diagnostics}
Table~\ref{tab:prefill-quality} covers 44 transitions from four OpenWAM
\texttt{adjust\_bottle} trajectories. Each starts with full initialization,
followed by 11 updates eight environment steps apart. Recursive variants
carry cache history and use their own feature anchors. An age cap of two
permits at most two consecutive reuses. Errors are measured against dense
FFN computation over the normalized $32\times20$ active action chunk.
At 64 refreshed positions, an age cap increases mean MAE from 0.002571
to 0.004006; increasing refresh to 96 with the cap gives 0.001596.

\begin{table}[!ht]
\normalsize
\caption{\textbf{Controlled OpenWAM token reuse in Prefill.} All 44 transitions, $N=2$; one L0-feature mask per observation. Mean and worst-observation MAE and maximum component error use normalized active actions. Future-video and action branches remain dense.}
\label{tab:prefill-quality}
\centering
\small
\begin{tabular*}{\linewidth}{@{\extracolsep{\fill}}lrrrr@{}}
\toprule
History / selector & Refresh & Mean MAE & Worst MAE & Max. error \\
\midrule
Dense previous history / L0 & 64/120 & 0.002722 & 0.022480 & 0.140625 \\
Recursive / L0 & 64/120 & 0.002571 & 0.011384 & 0.269531 \\
Recursive / L0 + age cap & 64/120 & 0.004006 & 0.035773 & 0.473633 \\
Recursive / L0 + RGB + age cap & 64/120 & 0.003292 & 0.016858 & 0.273438 \\
Recursive / L0 + age cap & 96/120 & 0.001596 & 0.012808 & 0.117188 \\
\bottomrule
\end{tabular*}
\end{table}

\clearpage
\section{Motion-Adaptive Refinement}
\label{app:rr-adaptive}\label{sec:adaptive-evaluation}\label{app:motion-motivation}

\subsection{Sampler and motion score}

\paragraph{Shared-prefix continuation.}
We integrate the predicted velocity field using Euler updates, following
the flow-matching formulation~\citep{lipman2023flowmatching}.
For updates $i=0,\ldots,9$, let $h_i=\operatorname{Embed}(x_i,s_i,c)$
be the current latent embedding, with timestep $s_i$ and conditioning $c$.
Writing $B$ for the full Transformer, we cache its input-to-output residual
\begin{equation}
 C_i=\begin{cases}
 B(h_i,s_i,c)-h_i, & i\in\mathcal E,\\
 C_{i-1}, & i\notin\mathcal E.
 \end{cases}
 \label{eq:transformer-residual-cache}
\end{equation}
The output head and Euler update are evaluated at every step:
\begin{equation}
 \widetilde v_i=\operatorname{Head}(h_i+C_i,s_i),
 \qquad x_{i+1}=x_i+\Delta t_i\widetilde v_i.
 \label{eq:cached-feature-euler}
\end{equation}
The cache is initialized at update 0 of each request. The 2F schedule uses
$\mathcal E_2=\{0,3\}$ and the 4F schedule uses
$\mathcal E_4=\{0,3,6,9\}$; the 10F reference refreshes at every update.
NFE counts full Transformer evaluations. All schedules share ten Euler updates
and fixed observation and instruction conditioning within a request.

After the second Transformer refresh at update 3, we decode a coarse action
prediction and measure its amplitude using
Equation~\ref{eq:adaptive-continuation}. If refinement is selected,
the runtime continues with additional Transformer refreshes at updates 6
and 9, retaining the same initial noise and conditioning. OpenWAM caches
the residuals of the future-video and action Transformers; FastWAM caches the action Transformer residual.

\paragraph{Coordinates and threshold.}
In the Cartesian-action probes, displacement is the maximum Euclidean
distance from current end-effector XYZ across 32 predicted positions,
using the training action statistics to restore metres. The joint-action
FastWAM runtime obtains XYZ by forward kinematics before evaluating
displacement. The recorded RoboTwin motion probes
use a fixed 0.08\,m threshold; rotation and gripper coordinates do not
enter this score.

\subsection{Paired prediction comparisons}

\begin{table}[!htbp]

\caption{\textbf{Motion refinement ablation.}
Means over 93 paired observations; green $\checkmark$/red $\times$:
enabled/disabled. Refine adds two evaluations; the motion gate selects
requests. XYZ deviations are relative to 10F; absolute values are in mm.}
\label{tab:motion-consistency}
\begin{center}
\begin{minipage}{0.76\linewidth}
\begingroup%
\normalsize%
\setlength{\tabcolsep}{0pt}%
\renewcommand{\arraystretch}{1.25}%
\newcommand{\motionon}{\textcolor[HTML]{39896D}{$\checkmark$}}%
\newcommand{\motionoff}{\textcolor[HTML]{BB655A}{$\times$}}%
\small
\begin{tabular}{@{}p{0.205\linewidth}>{\centering\arraybackslash}p{0.14\linewidth}>{\centering\arraybackslash}p{0.135\linewidth}>{\centering\arraybackslash}p{0.14\linewidth}>{\centering\arraybackslash}p{0.185\linewidth}>{\centering\arraybackslash}p{0.195\linewidth}@{}}
\toprule
\multirow{2}{*}{Policy} & \multicolumn{2}{c}{Components} & \multirow{2}{*}{\shortstack{Mean\\NFE}}
& \multicolumn{2}{c}{XYZ deviation $\downarrow$} \\
\cmidrule(lr){2-3}\cmidrule(l){5-6}
& Refine & \shortstack{Motion\\gate} & & \shortstack{Abs.\\(mm)} & Relative \\
\midrule
Fixed 2F & \motionoff & \motionoff & 2.00 & 7.65 & 0.0869 \\
Fixed 4F & \motionon & \motionoff & 4.00 & 3.68 & 0.0280 \\
\rowcolor{tablehighlight}
\textbf{Ours} & \motionon & \motionon & \textbf{2.65} & \textbf{6.48} & \textbf{0.0468} \\
\bottomrule
\end{tabular}%
\endgroup%

\end{minipage}
\end{center}
\end{table}

Deviation is the mean Euclidean XYZ error over both arms and the first
24 predicted steps. Relative deviation divides this error by the largest
10F displacement across those steps and arms, with a 0.02\,m floor.
Across 93 paired observations from one trajectory each of
\texttt{stack\_bowls\_two}, \texttt{blocks\_ranking\_size}, and
\texttt{put\_bottles\_dustbin} (18, 49, and 26 requests), we compare 2F
and 4F with the same model's 10F output. Table~\ref{tab:motion-consistency}
compares the fixed and adaptive budgets, weighting requests equally.
The gate selects 4F for 30 of 93 requests at a coarse 2F displacement
threshold of 0.08\,m. For these 30 small-motion requests,
mean relative deviation is 0.164 with 2F and 0.040 with 4F.
For the other 63 requests, the corresponding means are 0.050 and 0.022.
The absolute mean reductions are 3.62\,mm and 4.14\,mm, respectively.
The larger relative reduction in the small-motion group partly reflects
its smaller normalization denominator; its absolute 2F deviation is also lower.

\clearpage
\begin{figure}[!htbp]
\centering
\includegraphics[width=\linewidth]{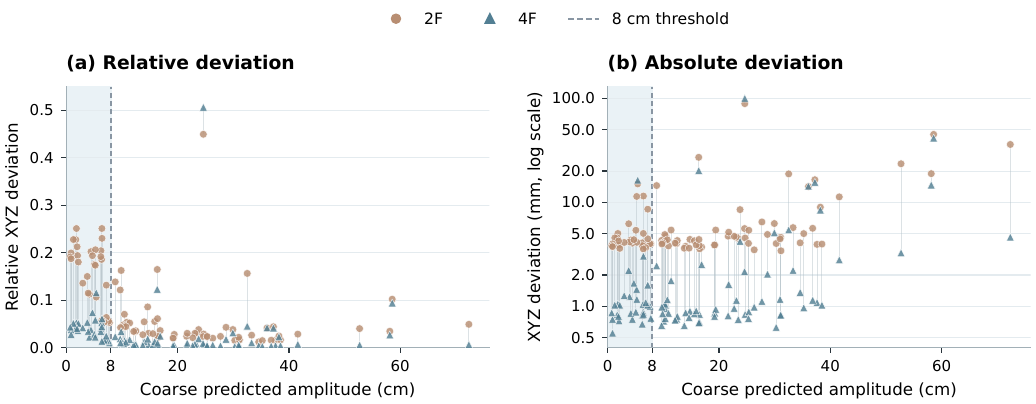}
\caption{\textbf{Prediction deviations over all 93 paired observations.}
Each request contributes a 2F and a 4F point at the same coarse predicted
displacement, joined by a faint vertical line. Both compare with the
same-input 10F prediction over the first 24 XYZ steps. Left: deviation
normalized by the 10F amplitude, with a 0.02\,m floor. Right: absolute
mean XYZ deviation on a logarithmic scale, in mm. The dashed vertical
line marks the 8\,cm motion-gate threshold.}
\label{fig:appendix-motion-distribution}
\end{figure}

Figure~\ref{fig:appendix-motion-distribution} separates relative and
absolute error. Small-motion requests show larger relative deviations
under 2F, while the absolute-error view reveals both the role of
normalization and variation across requests. Refinement reduces deviation
from 10F on 90 of the 93 observations and increases it on the remaining three.

\begin{table}[!htbp]
\normalsize
\caption{\textbf{Fixed-budget action deviations on FastWAM.}
Executed-prefix normalized MAE against the same-observation 10F
prediction over 1,798 observations.}
\label{tab:adaptive-evidence}
\begin{center}
\small
\begin{tabular*}{\linewidth}{@{\extracolsep{\fill}}lrrrr@{}}
\toprule
Evaluation positions & Mean & Median & P95 & Max \\
\midrule
$\{0,3\}$ & 0.007304 & 0.006876 & 0.010497 & 0.027701 \\
$\{0,3,6,9\}$ & 0.001224 & 0.001186 & 0.001713 & 0.005981 \\
\bottomrule
\end{tabular*}
\end{center}
\end{table}

A separate FastWAM replay covers 1,798 observations
(Table~\ref{tab:adaptive-evidence}). Four evaluations reduce mean
executed-prefix action MAE against the 10F reference from 0.007304 to 0.001224,
and P95 from 0.010497 to 0.001713. Fixed-budget comparisons share the
observation, initial noise, integration schedule, and action normalization.

\subsection{Visualization settings}
Figure~\ref{fig:motivation}a displays requests 1--25 of the
\texttt{put\_bottles\_dustbin} episode; all 26 requests enter the paired
analysis. Predicted speed is end-effector displacement per control step.
The display smooths 608 samples with a Gaussian ($\sigma=4$ steps,
radius 12, reflected endpoints), then retains every fourth sample.
The two deviation strips use 2F/4F predictions against 10F, with the
relative metric above and a common range of 0--0.25. Images are recorded
request inputs. Panel b uses schematic paths and task-tolerance radii.

Figure~\ref{fig:dynamic-allocation} uses all 50 requests from one
OpenWAM \texttt{blocks\_ranking\_size} rollout. At the 8\,cm threshold,
30 requests use 2F and 20 use 4F. Token capacities are 16, 80, and 120,
with full initialization. The episode ends in task failure.

\end{document}